\documentclass[10pt,twocolumn,letterpaper]{article}

\usepackage{iccv}
\usepackage{times}
\usepackage{epsfig}
\usepackage{graphicx}
\usepackage{amsmath}
\usepackage{amssymb}
\usepackage{bm}
\usepackage{subfigure}
\usepackage{bbding}
\usepackage{booktabs}
\usepackage{pifont}
\usepackage{multirow}
\usepackage[accsupp]{axessibility}
\newcommand\norm[1]{\lVert#1\rVert}
\newcommand{\xmark}{\ding{55}}%

\usepackage[breaklinks=true,bookmarks=false]{hyperref}

\iccvfinalcopy 

\def\iccvPaperID{6192} 

\ificcvfinal\fi

\begin{document}

\title{Semantics Meets Temporal Correspondence: Self-supervised \\Object-centric Learning in Videos}

\author {
    Rui Qian\textsuperscript{\rm 1} \quad
    Shuangrui Ding\textsuperscript{\rm 1} \quad Xian Liu\textsuperscript{\rm 1}  \quad Dahua Lin\textsuperscript{\rm 1,2}\thanks{Corresponding author. Email: dhlin@ie.cuhk.edu.hk.}\\
    \textsuperscript{\rm 1}The Chinese University of Hong Kong \quad 
    \textsuperscript{\rm 2}Shanghai Artificial Intelligence Laboratory\\
    {\tt\small \{qr021,ds023,lx021,dhlin\}@ie.cuhk.edu.hk}
}

\maketitle
\ificcvfinal\thispagestyle{empty}\fi

\begin{abstract}
Self-supervised methods have shown remarkable progress in learning high-level semantics and low-level temporal correspondence. Building on these results, we take one step further and explore the possibility of integrating these two features to enhance object-centric representations. Our preliminary experiments indicate that query slot attention can extract different semantic components from the RGB feature map, while random sampling based slot attention can exploit temporal correspondence cues between frames to assist instance identification. Motivated by this, we propose a novel semantic-aware masked slot attention on top of the fused semantic features and correspondence maps. It comprises two slot attention stages with a set of shared learnable Gaussian distributions. In the first stage, we use the mean vectors as slot initialization to decompose potential semantics and generate semantic segmentation masks through iterative attention. In the second stage, for each semantics, we randomly sample slots from the corresponding Gaussian distribution and perform masked feature aggregation within the semantic area to exploit temporal correspondence patterns for instance identification. We adopt semantic- and instance-level temporal consistency as self-supervision to encourage temporally coherent object-centric representations. Our model effectively identifies multiple object instances with semantic structure, reaching promising results on unsupervised video object discovery. Furthermore, we achieve state-of-the-art performance on dense label propagation tasks, demonstrating the potential for object-centric analysis. The code is released at \url{https://github.com/shvdiwnkozbw/SMTC}.
\end{abstract}

\begin{figure}
    \centering
    \subfigure[Query slot attention on semantic feature.]{
    \includegraphics[width=0.45\linewidth]{fig/camel.jpg}
    \includegraphics[width=0.45\linewidth]{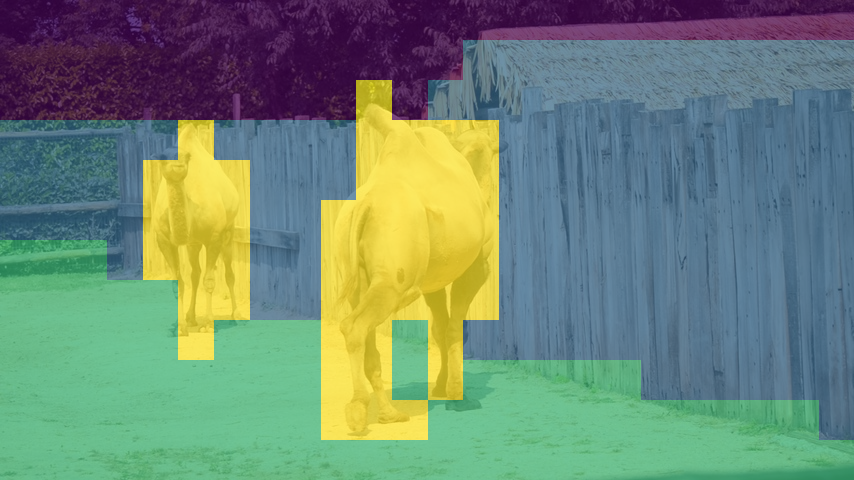}
    \label{semantic}
    }\\
    \subfigure[Random sampling slot attention on correspondence map.]{
    \includegraphics[width=0.45\linewidth]{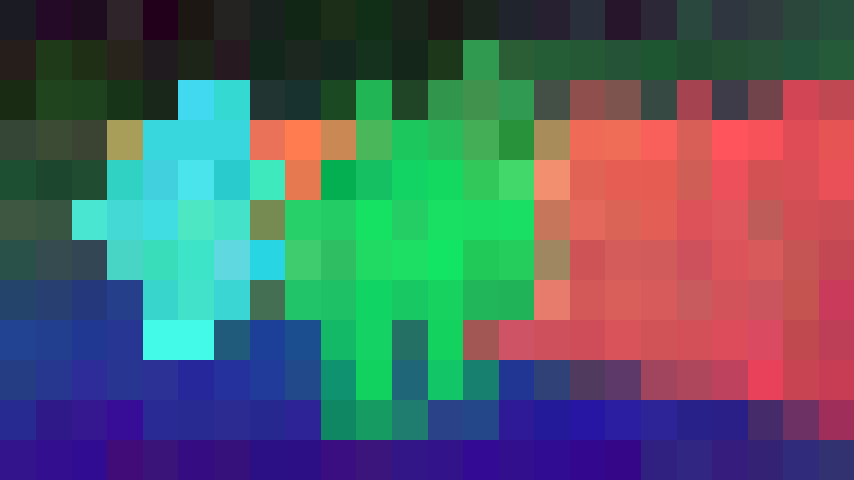}
    \includegraphics[width=0.45\linewidth]{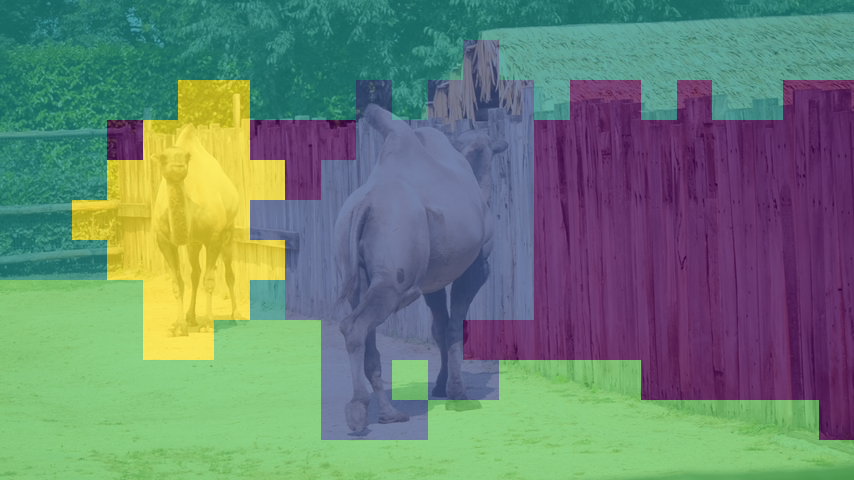}
    \label{correspondence}
    }
    \caption{Fig.~\ref{semantic} presents the results of query slot attention on top of the RGB feature map. It successfully decomposes different semantics, e.g., camels and fence. Fig.~\ref{correspondence} visualizes the correspondence map after PCA dimensionality reduction, showing that different instances have different correspondence patterns. And the slot attention with random sampling coarsely distinguishes two camels with some redundant borders. Best viewed in color.}
    \label{slotattention}
\end{figure}

\section{Introduction}
As one of the fundamental cognitive capabilities, human beings easily distinguish different objects, establish visual correspondence and perform object-centric analysis from temporally continuous observations. This ability can be attributed to two indispensable visual mechanisms: high-level semantic discrimination as well as low-level temporal correspondence, which enable humans to effectively understand and interact with the world. 

Motivated by this, computer vision researchers equip machines with these capacities to enhance object-centric perception~\cite{wang2013learning,long2015fully,han2017scnet}. To achieve this goal, early works rely on human annotations or weak supervision to perceive object semantics~\cite{long2015fully,ng2018actionflownet,carreira2017quo}, identify geometric positions~\cite{he2017mask,ren2015faster,girshick2015fast,zhu2017flow}, and establish temporal correspondence~\cite{held2016learning,wang2013learning,valmadre2017end,voigtlaender2017online}, but their generalization ability is limited. Recently, there emerge a host of fully unsupervised methods to learn robust representations for semantic discrimination~\cite{chen2020simple,he2020momentum,grill2020bootstrap,qian2021spatiotemporal,feichtenhofer2021large,he2022masked,caron2021emerging} or spatio-temporal correspondence~\cite{jabri2020space,wang2019learning,xu2021rethinking,li2019joint,lai2019self,hu2022semantic}, which achieve promising performance. Given this encouraging result, we naturally come up with a question: Is it possible to jointly leverage the semantics and correspondence to discover object instances and distill object-centric representations without human annotations?

Regarding this problem, our intuition is that the high-level semantics delineates meaningful foreground areas in a top-down manner, while when looking into more frames, the low-level correspondence temporally associates coherent objects and separates individual instances in a bottom-up fashion. For instance, in a football scene, the semantic cue differentiates the foreground that includes several players, while the temporal correspondence links distinct players through dynamic movements and geometric relationships. These two aspects collectively contribute to object-centric representations. Unfortunately, most of the existing works only concentrate on one of these features. \cite{chen2020simple,he2020momentum,grill2020bootstrap,qian2021spatiotemporal,wang2021dense} succeed in developing high-level semantics, but this abstract semantics alone is insufficient to distinguish instances. Whereas, \cite{jabri2020space,wang2019learning,vondrick2018tracking,li2019joint} excel in detailed correspondence, but lack semantic structure and result in redundancy and ambiguity.

In this paper, we propose a new architecture, \textit{Semantics Meets Temporal Correspondence (SMTC)}, to jointly leverage semantics and temporal correspondence to distill object-centric representations from RGB sequences. Specifically, we first extract frame-wise visual features as the semantic representation. Then we calculate dense feature correlations between adjacent frames as the correspondence map which encodes temporal relationships. To mine the object-centric knowledge, we take inspiration from~\cite{locatello2020object,yang2021self,kipf2021conditional,elsayed2022savi++}, and investigate using different formulations of slot attention on them. The preliminary experiments show that the original slot attention with random sampling on RGB feature map suffers from complex scene components in real-world videos~\cite{locatello2020object,seitzer2022bridging}, but the revised query slot attention~\cite{yang2021self,jia2022unsupervised} can decompose different semantic components as shown in Fig.~\ref{semantic}. As for the correspondence map, different objects present diverse temporal correspondence patterns. Comparing to semantic features, these patterns reveal low-level geometric relationships, which are comparatively simple but vary with specific scenes. Hence, query slot attention fails but the random sampling based formulation performs surprisingly well as shown in Fig.~\ref{correspondence}, coarsely separating different object areas with some redundant borders.

Motivated by this, we develop semantic-aware masked slot attention, which comprises a set of Gaussian distributions with learnable mean and standard deviation vectors, on top of the fused semantic and correspondence representations. The intuition is that the mean vectors could represent potential semantic centers, which act similarly to the query slot attention to separate semantic components.
While the deviation vectors introduce perturbations around the semantic centers to capture distinct temporal correspondence patterns of different instances.
Technically, we formulate two slot attention stages to achieve this goal. Firstly, we use the mean vectors as slot initialization to generate semantic segmentation masks. Secondly, for each semantics, we randomly sample slot vectors from the Gaussian distribution, then perform iterative attention and masked aggregation within the corresponding semantic mask area to distinguish instances.
We enforce temporal consistency on the semantic masks as well as object instance slots to enhance temporal coherency and refine object-centric representations. Comparing with existing works on object-centric learning in videos~\cite{elsayed2022savi++,kipf2021conditional,yang2021self,xie2022segmenting}, our model is free of pre-computed motion or depth prior, and explicitly identifies multiple objects with semantic structure.


In summary, our contributions are: (1) We propose a novel self-supervised architecture that unifies semantic discrimination and temporal correspondence to distill object-centric representations in videos. (2) We demonstrate that simple feature correlation can effectively represent temporal correspondence cues when used in conjunction with semantic features. Building on this observation, we develop semantic-aware masked slot attention, which operates on fused visual features and correspondence maps, to distinguish multiple object instances with semantic structure without relying on motion or depth priors.
(3) We achieve promising results on unsupervised object discovery in both single and multiple object scenarios, and reach state-of-the-art performance on label propagation tasks, demonstrating that we learn discriminative and temporally consistent object-centric representations.


\begin{figure*}
    \centering
    \includegraphics[width=\linewidth]{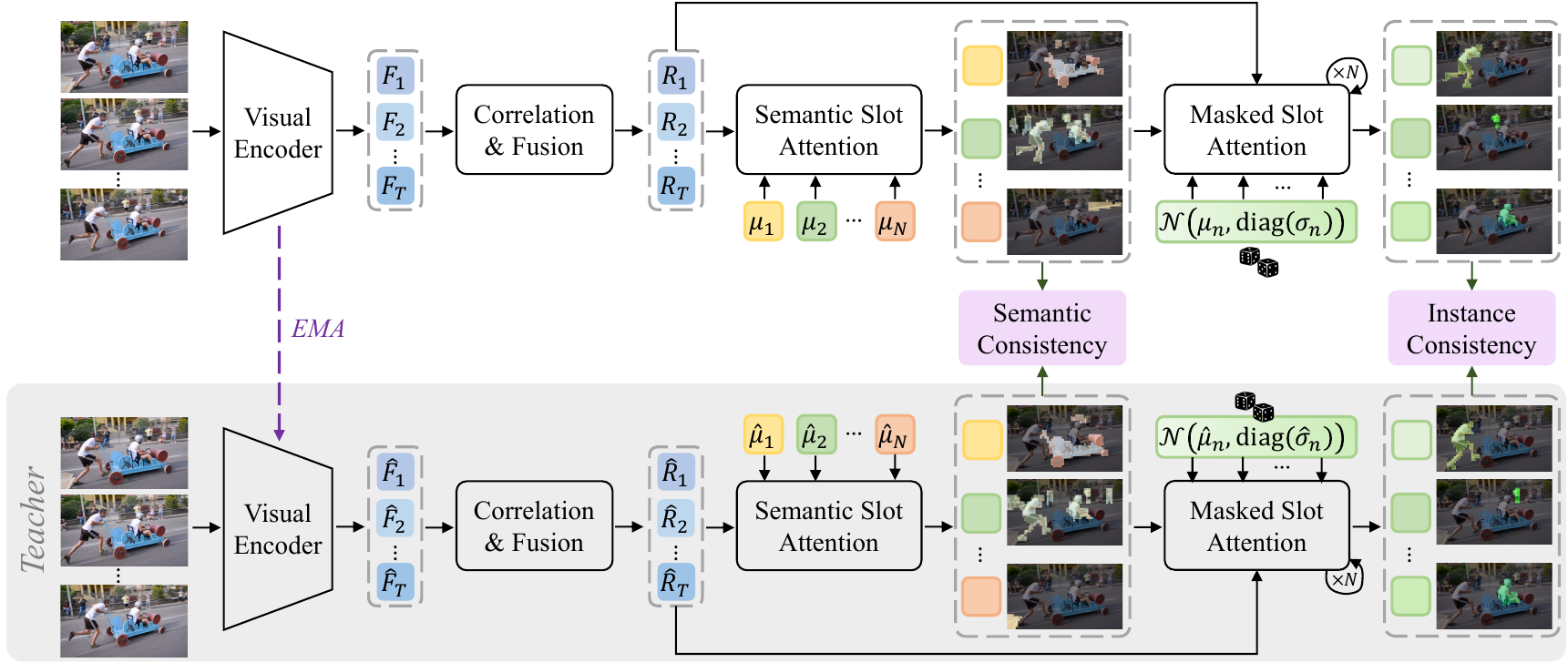}
    \caption{An overview of our framework. We first extract frame-wise features and calculate dense feature correlation, then fuse them to pass through semantic-aware masked slot attention, which comprises two slot attention stages with $N$ shared learnable Gaussian distributions. In the first semantic slot attention stage, the mean vectors serve as slot initialization to generate a set of segmentation masks for semantic decomposition. In the second masked slot attention stage, which runs on $N$ semantics in parallel, we randomly sample slots from the Gaussian distribution of each semantics and perform masked feature aggregation within the semantic area to identify distinct instances. We enforce semantic and instance temporal consistency to train the architecture in a teacher-student manner, with the teacher marked in \textcolor[rgb]{0.8, 0.8, 0.8}{gray}.}
    \label{framework}
\end{figure*}
\section{Related Work}
\noindent\textbf{Unsupervised Object Discovery} is an essential process to formulate object-centric representations, which aims to identify objects without human annotations. There exist a series of works focusing on this problem~\cite{locatello2020object,greff2019multi,emami2021efficient,engelcke2019genesis,engelcke2021genesis,burgess2019monet}. Typically, \cite{locatello2020object} develops slot attention to iteratively update latent object representations from random initializations but it has difficulty scaling to complex real-world scenes. To tackle this challenge, \cite{seitzer2022bridging} employs feature reconstruction as objective to reduce redundancy, \cite{jia2022unsupervised} develops query slot initialization to encode visual concepts. Further, a line of works extend object discovery to video domain~\cite{kipf2021conditional,kabra2021simone,crawford2020exploiting,besbinar2021self,yang2021self}. Most of them build on slot attention architecture, and adopt extra pre-computed priors, e.g., optical flow~\cite{yang2021self,xie2022segmenting,choudhury2022guess,ding2022mod}, depth~\cite{elsayed2022savi++}, geometric positions~\cite{kipf2021conditional}, as input or supervision to assist object discovery. In our work, we find that the simple feature correlation provides informative temporal cues. And we develop semantic-aware masked slot attention to identify object instances with semantic structure without resorting to extra priors.

\noindent\textbf{Self-supervised Representation Learning} aims to learn robust representation without human annotations. Early works design various pretext tasks to generate pseudo labels as self-supervision~\cite{kim2018learning,gidaris2018unsupervised,jing2018self,caron2018deep,alwassel2020self,misra2016shuffle}. Later, contrastive learning with instance discrimination revolutionized the field~\cite{he2020momentum,chen2020simple,qian2021spatiotemporal,feichtenhofer2021large,han2019video,wu2018unsupervised,oord2018representation,henaff2020data,caron2020unsupervised,ding2022dual,bachman2019learning,qian2022static,han2019video,han2020self,qian2021enhancing,ding2022motion,ding2023prune}. These works generally contribute to general semantics but lack the ability to distill object-centric representations. To delve into this problem, \cite{wang2021dense,xie2021propagate} preserve spatial sensitivity and perform contrastive learning on dense features. \cite{xu2021rethinking,caron2021emerging} point out that semantic pre-training implicitly encodes object correspondence.  While in our work, we learn to explicitly identify different objects with instance-level consistency to progressively refine object-centric representations. 

\noindent\textbf{Self-supervised Correspondence} is a fundamental problem in computer vision~\cite{li2019joint,vondrick2018tracking,wang2019learning,wang2017transitive}. Typical works rely on low-level statistics like color to generate the dense correspondence flow~\cite{lai2020mast,lai2019self,vondrick2018tracking}, or establish temporal cycles to track image regions~\cite{wang2019learning,jabri2020space,bian2022learning}. Further, \cite{hu2022semantic,xu2021rethinking,caron2021emerging} incorporate correspondence learning with semantic discrimination, but they either regard correspondence as a side effect of semantic discrimination~\cite{xu2021rethinking,caron2021emerging} or independently learn two pathway features with late fusion~\cite{hu2022semantic}. Among these works, high-level semantics is not well utilized, and there exists high redundancy due to large background areas. In contrast, our method explicitly separates semantic components to alleviate redundancy and formulate instance-level correspondence from an object-centric perspective.

\section{Method}
Our framework is shown in Fig.~\ref{framework}, which adopts a teacher-student structure similar to recent self-supervised methods~\cite{he2020momentum,caron2021emerging,grill2020bootstrap}. Specifically, we first extract frame-wise features from an RGB sequence, and calculate dense feature correlation as temporal correspondence representations. After that, we fuse them to pass through semantic-aware masked slot attention to separate multiple object instances with semantic structure, and enforce temporal consistency to refine the object-centric representations.

\subsection{Feature Encoding}
Given a RGB sequence $v=\{I_1,I_2,...,I_T\}$, we employ a shared visual encoder $f$ to extract frame-wise features:
\begin{align}
    F_t = f(I_t) \in\mathbb{R}^{HW\times D},
\end{align}
where $H,W,D$ respectively denotes height, width and channel dimension, $t$ indicates frame index. For simplicity, we omit the subscript indicating student or teacher pathway, and use $\hat{F}$ to denote teacher outputs. Intuitively, $F_t$ contains appearance and semantic cues in the input sequence, and the next step is to formulate an effective representation for the temporal correspondence cues. To do this, inspired by the cost volume in motion estimation~\cite{dosovitskiy2015flownet,sun2018pwc,teed2020raft}, we calculate dense feature correlation to generate the temporal correspondence map for each timestamp:
\begin{align}
    C_{tj} = F_tF_j^T \in\mathbb{R}^{HW\times HW},
\end{align}
where we randomly sample one frame index $j\neq t$ to compute feature correlation. The channel activations of $C_{tj}$ encode low-level geometric correlations between frame $t$ and $j$, which indicates potential partitions of different objects as illustrated in Fig.~\ref{correspondence}. 

To jointly leverage the semantic and correspondence cues, we employ two linear transformation heads, $h_f$ and $h_c$, to respectively project $F_t$ and $C_{tj}$ to a shared $D$-dimensional embedding space, which are then fused with element-wise summation:
\begin{align}
    R_t = h_f(F_t) + h_c(C_{tj}) \in\mathbb{R}^{HW\times D}.
\end{align}
In this way, $R_t$ simultaneously encodes semantic and temporal correspondence information, serving as an intermediate feature for object-centric analysis.

\begin{table}[]
    \centering
    \begin{tabular}{ccc}
    \toprule
        Feature & Query & Random \\
        \midrule
        RGB & 58.1 & 38.4 \\
        Correlation & 21.5 & 56.3 \\
        \bottomrule
    \end{tabular}
    \caption{Preliminary results on RGB feature map and feature correlation. `Query' (`Random') denotes slot attention with query (random sampling) initialization. We report IoU on DAVIS-2016.}
    \label{preliminary}
\end{table}

\subsection{Preliminary Experiments with Slot Attention}
Given these representations, the next step is to formulate effective ways to identify individual objects. To do this, we take inspiration from slot attention~\cite{locatello2020object,kipf2021conditional,yang2021self,jia2022unsupervised}, which employs a set of slot vectors to iteratively attend to specific objects. Generally, there are two alternatives for slot initialization, random sampling from a Gaussian distribution with learnable parameters $\mu,\sigma\in\mathbb{R}^D$~\cite{locatello2020object,kipf2021conditional} or directly inheriting from a set of learnable queries~\cite{jia2022unsupervised,yang2021self}.

To investigate the effectiveness of these two formulations, we conduct preliminary experiments on the semantic feature $F_t$ and correspondence map $C_{tj}$ respectively. We follow the settings in~\cite{seitzer2022bridging}, using a frozen DINO pre-trained ViT to extract visual feature and generate correspondence map, on top of which we perform two kinds of slot attention. We train and evaluate them on DAVIS-2016 dataset~\cite{perazzi2016benchmark}. As indicated by Fig.~\ref{slotattention} and Table.~\ref{preliminary}, we find that query slot attention decomposes distinct semantic components on $F_t$, while the original slot attention with random sampling performs much worse and struggles in complex real-world scenes. Conversely, query slot attention does not work on the correspondence map, and random sampling initialization enables the model to effectively utilize the temporal correspondence patterns in $C_{tj}$ to distinguish instances. An intuitive explanation into these observations is the RGB features contain rich semantics and there exist consistent patterns for similar objects. Slot attention with learnable queries can capture inter-sample semantic patterns to identify objects. While random initialization fails to memorize semantic patterns, thus resulting in worse performance. In contrast, correspondence features reveal inter-frame geometric correlations, shown in Fig.~\ref{correspondence}, which vary with specific scenes. There are no consistent patterns for similar objects, so the learnable queries fail but random initialization works well to capture correspondence cues. The complementarity between these two conditions motivates two-stage slot attention design to combine the attributes of query and random sampling initialization, thus better exploiting the semantic and correspondence cues in $R_t$.

\subsection{Semantic-aware Masked Slot Attention}
We propose semantic-aware masked slot attention, which comprises $N$ Gaussian distributions with learnable means $\bm{\mu}=\{\mu_1,\mu_2,...,\mu_N\}\in\mathbb{R}^{N\times D}$ and standard deviations $\bm{\sigma}=\{\sigma_1,\sigma_2,...,\sigma_N\}\in\mathbb{R}^{N\times D}$. With this design, the mean vectors act similarly to the learnable queries, each representing a potential semantic center. And the deviations introduce perturbations around semantic centers to model temporal correspondence patterns. To achieve this goal, we employ two stages of iterative attention with shared learnable parameters. First, we use the mean vectors as slot initialization to decompose different semantics in the form of segmentation. Second, we randomly sample slots from each Gaussian distribution $(\mu_n,\text{diag}(\sigma_n))$, and perform masked aggregation within the corresponding semantic area to identify instances. The process is detailed as follows.

\noindent\textbf{Semantic Decomposition.} In the first stage, we directly use $\bm{\mu}$ without randomness as slot initialization $S=\bm{\mu}\in\mathbb{R}^{N\times D}$. Then following the standard slot attention procedure~\cite{locatello2020object}, we employ three linear transformation heads to map $S$ and $R_t$ into $Q\in\mathbb{R}^{N\times D}, K,V\in\mathbb{R}^{HW\times D}$, and iteratively calculate attention score and update slot representations. Mathematically, we formulate each iteration as:
\begin{align}
    &M := \text{softmax}(\frac{1}{\sqrt{D}}QK^T, 0)\in\mathbb{R}^{N\times HW}, \label{attention} \\
    &A[n,i] := \frac{M[n,i]}{\sum_l M[n,l]+\epsilon},\text{ } A\in\mathbb{R}^{N\times HW}, \label{normalize}\\
    &S := \text{GRU}(\text{input}=AV,\text{states}=S).
\end{align}
Note that the slot attention weight $M$ is normalized along the \texttt{query} axis, and the weighted mean coefficient $A$ is employed to aggregate the \texttt{value} to update the slots. This mechanism induces competition among slots and enforce different slots to take over different semantic features. We iterate this process for three times, and take the slot attention weight (slot vectors) of the final iteration as the segmentation mask (representation) for different semantic components. To avoid ambiguity in presentation, we denote them as $\mathcal{M}\in\mathbb{R}^{N\times HW}$ and $\mathcal{S}\in\mathbb{R}^{N\times D}$, where $\mathcal{S}[n]$ presents $n$-th semantic center representation, $\mathcal{M}[n]$ indicates the probability that each pixel belongs to $n$-th semantics.

\noindent\textbf{Instance Identification.} Within each semantics, there could exist multiple object instances. The Gaussian distributions we introduce help to distinguish them with temporal correspondence cues. Specifically, in the second stage, for each semantics, we randomly sample slots from the Gaussian distribution, using the perturbations around the semantic center to capture distinct temporal correspondence patterns. For simplicity, we take the $n$-th semantics as an example. First, we sample $P$ vectors, $S\sim\mathcal{N}(\mu_n,\text{diag}(\sigma_n))\in\mathbb{R}^{P\times D}$, as slot initialization to represent $P$ potential instances. Next, we follow the first stage formulation to respectively project $S$ and $R_t$ into $Q\in\mathbb{R}^{P\times D}$ and $K,V\in\mathbb{R}^{HW\times D}$, and obtain the attention score $M\in\mathbb{R}^{P\times HW}$. To preserve the semantic attributes and discriminate correspondence cues, we perform masked feature aggregation within the semantic mask areas. The only mathematical difference in this process is the weighted mean coefficient computation:
\begin{align}
\begin{split}
    &A[p,i] := \frac{M[p,i]\widetilde{\mathcal{M}}[n,i]}{\sum_l M[p,l]\widetilde{\mathcal{M}}[n,l]+\epsilon},\text{ } A\in\mathbb{R}^{P\times HW},\\
    &\widetilde{\mathcal{M}}[n,i] = \text{binarize}(\mathcal{M}[n,i],\tau),\text{ } \mathcal{M}\in\mathbb{R}^{N\times HW},
\end{split}
\end{align}
where we use the $n$-th semantic mask $\widetilde{\mathcal{M}}[n]\in\{0,1\}^{HW}$ binarized with threshold $\tau$ to constrain the model to only aggregate the visual features related to $n$-th semantics. And we empirically find a comparatively large threshold $\tau=0.5$ helps to preserve object regions with clear semantics and filter out background. The masked slot attention for all semantics follows the same rule and is calculated in parallel. Similarly, we iterate the process for three times, use the final slot attention weight as the instance segmentation masks, and take the final slot vectors as the object-centric representations$\mathcal{O}\in\mathbb{R}^{N\times P\times D}$, where $\mathcal{O}[n,p]\in\mathbb{R}^D$ denotes the $p$-th potential object instance of $n$-th semantics. In this way, our semantic-aware masked slot attention explicitly segments multiple instances with semantic structure. 


\subsection{Training}
We use temporal consistency as self-supervision to optimize semantic masks $\mathcal{M}$ and instance representations $\mathcal{O}$.

\noindent\textbf{Dense Semantic Alignment.} Given the semantic mask of each timestamp $\mathcal{M}_t\in\mathbb{R}^{N\times HW}$, as well as $\hat{\mathcal{M}}_t$ from teacher pathway, we aim to align the spatially dense semantic distributions across time. To achieve this, the first step is to determine the corresponding patches between different frames. Utilizing the feature correlation $C_{tj}\in\mathbb{R}^{HW\times HW}$, which indicates dense feature similarity between timestamps $t,j$, we can infer patch-level correspondence from this cue with optimal transport~\cite{caron2020unsupervised,asano2019self,liu2020self,cuturi2013sinkhorn}. Formally, we take $-\hat{C}_{tj}$ from teacher pathway as cost matrix, and solve the optimal transport strategy $\pi^*_{tj}\in\mathbb{R}^{HW\times HW}$ between two uniform distributions to indicate patch correspondence:
\begin{align}
\begin{split}
    \min_{\pi_{tj}} &\sum_{u=1}^{HW}\sum_{v=1}^{HW}-\hat{C}_{tj}[u,v]\pi_{tj}[u,v]\\
    \text{s.t.} &\sum_{v=1}^{HW}\pi_{tj}[\cdot,v]=\frac{1}{HW}\bm{1}^{HW},\\
    &\sum_{u=1}^{HW}\pi_{tj}[u,\cdot]=\frac{1}{HW}\bm{1}^{HW},\\
    &\pi_{tj}[u,v]\geq 0\quad u,v\in\{1,2,...,HW\},
\end{split}
\label{optimal}
\end{align}
where $u,v$ denotes spatial index, $\bm{1}^{HW}$ is a $HW$-dimensional vector of all ones. Note that it is feasible to adopt other marginal distribution formulations with prior knowledge, e.g., class-agnostic activation map~\cite{baek2020psynet}, to facilitate training, with details discussed in Supplementary Material. We employ Sinkhorn-Knopp algorithm~\cite{cuturi2013sinkhorn} to obtain the optimal transport matrix $\pi_{tj}^*$, and formulate the dense semantic alignment objective in the form of cross entropy:
\begin{align}
    \mathcal{L}_{sem} = -\sum_{t,j=1\atop t\neq j}^T\sum_{u,v=1}^{HW}\sum_{n=1}^N\pi_{tj}^*[u,v]\hat{\mathcal{M}}_j[n,v]\log \mathcal{M}_t[n,u].
\end{align}
By minimizing the weighted sum of cross entropy, we achieve temporally consistent semantic distributions among corresponding spatial areas.

\noindent\textbf{Semantic Mask Regularization.} To encourage the semantic centers to emphasize different visual contents and cover diverse semantics, we apply a simple regularization to $\mathcal{M}_t$:
\begin{align}
    \mathcal{L}_{reg} = \sum_{t=1}^T\sum_{n,j=1\atop j\neq n}^N\frac{\mathcal{M}_t[n]^T\mathcal{M}_t[j]}{\norm{\mathcal{M}_t[n]}_2\norm{\mathcal{M}_t[j]}_2}.
\end{align}
This simple regularization suppresses the cosine similarity between different semantic masks to avoid collapse.

\noindent\textbf{Instance Representation Consistency.} Apart from the semantic distributions, it is necessary to ensure that the representations of each object instance are temporally consistent. Given the instance representations $\mathcal{O}_t\in\mathbb{R}^{N\times P\times HW}$, as well as $\hat{\mathcal{O}}_t$ from teacher pathway, we need to first match corresponding instances between different timestamps. For illustration, considering $n$-th semantics of time $t,j$, we adopt bipartite matching~\cite{carion2020end,cheng2021per} based on the cosine similarity between $\mathcal{O}_t[n]$ and $\hat{\mathcal{O}}_j[n]$ to generate one-to-one instance correspondence, with the matching function denoted as $\varepsilon(\cdot)$. In this way, $\mathcal{O}_t[n,p]$ and $\hat{\mathcal{O}}_j[n,\varepsilon(p)]$ are considered as a positive pair to be aligned. However, for each video, there exists absent semantics and object occlusion such that the number of visible object instances varies. Thus, there could be redundant slots not attending to objects, and it is crucial to filter out these invalid instance representations to reduce distractions. Mathematically, for timestamp $t$, we introduce $\mathcal{I}_t\in\{0,1\}^{N\times P}$ as the valid instance indicator. For the $p$-th instance of $n$-th semantics to be valid, two criteria must be met: (1) The ratio of the $n$-th semantic area is above threshold $\tau_1=0.2$ to filter out non-existing semantics; (2) The instance representation is close to the semantic center representation with cosine similarity larger than $\tau_2=0.5$ to exclude redundant slots. The conditions are formulated as:
\begin{align}
    \mathcal{I}_t[n,p] = 1 \Leftrightarrow
    \left\{
    \begin{aligned}
    &\frac{1}{HW}\sum_{u=1}^{HW}\widetilde{\mathcal{M}}_t[n,u] \geq \tau_1, \\
    &\cos(\mathcal{S}_t[n],\mathcal{O}_t[n,p]) \geq \tau_2.
    \end{aligned}
    \right.
    \label{criteria}
\end{align}
And $\hat{\mathcal{I}}_j\in\{0,1\}^{N\times P}$ is defined in the same manner. We use a margin loss to encourage object detail consistency over valid instance representations:
\begin{align}
\begin{split}
    \mathcal{L}_{obj} = &\sum_{t,j=1\atop t\neq j}^T\sum_{n=1}^N\sum_{p=1}^P\mathcal{I}_t[n,p]\{\\
    &\hat{\mathcal{I}}_j[n,\varepsilon(p)]\norm{\mathcal{O}_t[n,p]-\hat{\mathcal{O}}_j[n,\varepsilon(p)]}_2 \\
    & +\sum_{q=1\atop q\neq\varepsilon(p)}^P\text{relu}(\lambda-\norm{\mathcal{O}_t[n,p]-\hat{\mathcal{O}}_j[n,q]}_2)\}
\end{split}
\end{align}
where $\lambda$ is a margin hyper-parameter. Note that with the $\mathcal{I}_t$ and $\hat{\mathcal{I}}_j$ constraints, our formulation can handle object occlusions which lead to varying number of visible instances in different frames. By minimizing the margin loss, we encourage the model to distill discriminative and temporally consistent object-centric representations.

Overall, we take the summation of three objectives for training:
\begin{align}
    \mathcal{L} = \mathcal{L}_{sem} + \mathcal{L}_{obj} + \mathcal{L}_{reg}.
\end{align}
The student pathway $\theta$ is updated with gradient descent, and the teacher parameters $\hat{\theta}$ are updated with momentum as:
\begin{align}
    \hat{\theta} \leftarrow m\hat{\theta} + (1-m)\theta,
\end{align}
where $m$ is momentum coefficient set to 0.999 in default. This momentum update mechanism results in a slowly evolving teacher network, which progressively distills object knowledge, provides reliable self-supervision signals and enables us to train semantic-aware masked slot attention without relying on widely adopted reconstruction objective in existing works~\cite{locatello2020object,kipf2021conditional,yang2021self,xie2022segmenting,seitzer2022bridging,jia2022unsupervised}.

\section{Experiments}
\subsection{Dataset}
We train our model on YouTube-VOS~\cite{xu2018youtube}, a challenging video dataset that contains multiple object instances of distinct semantics in each video. We evaluate our method on two lines of tasks: (1) Unsupervised object segmentation on DAVIS-2016~\cite{perazzi2016benchmark}, SegTrack-v2~\cite{li2013video}, FMBS-59~\cite{ochs2013segmentation} and challenging multiple object segmentation on DAVIS-2017-Unsupervised~\cite{caelles20192019}. We respectively calculate the mean per frame intersection over union (IoU) and $\mathcal{J}\&\mathcal{F}$ on single and multiple object discovery benchmarks. (2) Lable propagation tasks including semi-supervised video object segmentation on DAVIS-2017~\cite{pont20172017}, human pose tracking on JHMDB~\cite{jhuang2013towards}, human part tracking on VIP~\cite{zhou2018adaptive}. We adopt the same settings as~\cite{jabri2020space,hu2022semantic} and report standard $\mathcal{J}$ and $\mathcal{F}$ score on DAVIS, probability of a correct pose (PCK) on JHMDB and mean intersection over union (mIoU) on VIP.

\subsection{Implementation Details}
We sample $T=4$ frames with frame rate $\text{FPS}=4$ as the input RGB sequence, where each frame is augmented with random crop, horizontal flip and color jitter, and finally resized to $256 \times 256$. We adopt ViT-Small/16~\cite{dosovitskiy2020image} and ResNet-50~\cite{he2016deep} as the visual encoder, specified in each experiment, to extract frame-wise features. Then for semantic-aware masked slot attention, we set the number of learnable Gaussian distributions to $N=16$, the number of instances of each semantics to $P=4$ in default and use $3$ iterations to update slot representations and attention maps.

In training, the visual encoder is initialized with self-supervised pre-trained weights from DINO ViT-Small/16~\cite{caron2021emerging} or MoCo ResNet-50~\cite{he2020momentum}. We use AdamW optimizer~\cite{loshchilov2018decoupled} with learning rate $2\times 10^{-4}$ for batch size $128$ to update student parameters, and the teacher pathway is updated in momentum. In inference, we employ the same evaluation protocol as~\cite{jabri2020space} to evaluate the temporal object correspondence performance on label propagation tasks. And for object discovery, we maintain the valid instances filtered with Eq.~\ref{criteria} as candidate objects. For single object evaluation, we merge all candidate objects as foreground areas to calculate IoU. For multiple object benchmark, we follow~\cite{caelles20192019} to match ground-truth and our predictions and report $\mathcal{J}\&\mathcal{F}$ score.

\subsection{Comparison with State-of-the-art}
\begin{table}[]
    \centering
    \begin{tabular}{lccccc}
    \toprule
        Model & RGB & Flow & DAVIS & ST-v2 & FBMS \\
        \midrule
        CIS~\cite{yang2019unsupervised} & \checkmark & \checkmark & 71.5 & 62.5 & 63.5 \\
        AMD~\cite{liu2021emergence} & \checkmark & \xmark & 57.8 & 57.0 & 47.5 \\
        DINO~\cite{caron2021emerging} & \checkmark & \xmark & 52.3 & 46.5 & 50.3 \\
        SIMO~\cite{lamdouar2021segmenting} & \xmark & \checkmark & 67.8 & 62.0 & - \\
        MG~\cite{yang2021self} & \xmark & \checkmark & 68.3 & 58.6 & 53.1 \\
        OCLR~\cite{xie2022segmenting} & \xmark & \checkmark & \textbf{72.1} & 67.6 & 65.4 \\
        GWM~\cite{choudhury2022guess} & \checkmark & \checkmark & 71.2 & 69.0 & 66.9 \\
        \midrule
        \textbf{SMTC} & \checkmark & \xmark & 71.8 & \textbf{69.3} & \textbf{68.4} \\
        \textbf{SMTC}$^\dagger$ & \checkmark & \xmark & 70.8 & 68.4 & 66.5 \\
        \bottomrule
    \end{tabular}
    \caption{Quantitative results on single object discovery. We compare per frame mean IoU on DAVIS-2016, SegTrack-v2 and FBMS-59 without any post-processing. SMTC$^\dagger$ denotes only with first slot attention stage for semantic decomposition in inference.}
    \label{discover}
\end{table}

\begin{table}[]
    \centering
    \begin{tabular}{lcccc}
    \toprule
        Model & Backbone & $\mathcal{J}\&\mathcal{F}$ & $\mathcal{J}$ & $\mathcal{F}$ \\
        \midrule
        DINOSAUR~\cite{seitzer2022bridging} & ViT-S/16 & 21.4 & 19.2 & 23.7 \\
        \textbf{SMTC} & ViT-S/16 & \textbf{40.5} & \textbf{36.4} & \textbf{44.6} \\
        \textbf{SMTC} & ResNet-50 & 39.0 & 35.5 & 42.6 \\
        \midrule
        \textcolor[rgb]{0.75, 0.75, 0.75}{RVOS}~\cite{ventura2019rvos} & \textcolor[rgb]{0.75, 0.75, 0.75}{ResNet-101} & \textcolor[rgb]{0.75, 0.75, 0.75}{41.2} & \textcolor[rgb]{0.75, 0.75, 0.75}{36.8} & \textcolor[rgb]{0.75, 0.75, 0.75}{45.7} \\
        \textcolor[rgb]{0.75, 0.75, 0.75}{ProReduce}~\cite{lin2021video} & \textcolor[rgb]{0.75, 0.75, 0.75}{ResNet-101} & \textcolor[rgb]{0.75, 0.75, 0.75}{68.3} & \textcolor[rgb]{0.75, 0.75, 0.75}{65.0} & \textcolor[rgb]{0.75, 0.75, 0.75}{71.6} \\
        \bottomrule
    \end{tabular}
    \caption{Quantitative results on multiple object discovery on DAVIS-2017-Unsupervised. \textcolor[rgb]{0.75, 0.75, 0.75}{Gray} denotes supervised training.}
    \label{multiple}
\end{table}

\begin{table*}[]
    \centering
    \begin{tabular}{lccccccc}
        \toprule
        {} & {} & \multicolumn{3}{c}{DAVIS} & \multicolumn{2}{c}{JHMDB} & VIP  \\
        \cmidrule(r){3-5}
        \cmidrule(r){6-7}
        \cmidrule(r){8-8}
        Model & Backbone & $\mathcal{J}\&\mathcal{F}$ & $\mathcal{J}$ & $\mathcal{F}$ & PCK@0.1 & PCK@0.2 & mIoU \\ 
        \midrule
        Supervised~\cite{he2016deep} & ResNet-50 & 66.0 & 63.7 & 68.4 & 59.2 & 78.3 & 39.5 \\
        MoCo~\cite{he2020momentum} & ResNet-50 & 65.4 & 63.2 & 67.6 & 60.4 & 79.3 & 36.1 \\
        VFS~\cite{xu2021rethinking} & ResNet-50 & 68.9 & 66.5 & 71.3 & 60.9 & 80.7 & 43.2 \\
        DINO~\cite{caron2021emerging} & ViT-S/16 & 63.8 & 61.5 & 66.1 & 45.4 & 75.2 & 37.9 \\
        DINO$^\dagger$~\cite{caron2021emerging} & ViT-S/16 & 67.5 & 65.2 & 69.9 & 53.0 & 79.5 & 39.1 \\
        \midrule
        TimeCycle~\cite{wang2019learning} & ResNet-50 & 40.7 & 41.9 & 39.4 & 57.7 & 78.5 & 28.9 \\
        UVC~\cite{li2019joint} & ResNet-50 & 56.3 & 54.5 & 58.1 & 56.5 & 76.6 & 34.2 \\
        MAST~\cite{lai2020mast} & ResNet-18 & 65.5 & 63.3 & 67.6 & - & - & - \\
        CRW~\cite{jabri2020space} & ResNet-18 & 67.6 & 64.8 & 70.2 & 58.8 & 80.3 & 37.6 \\
        SFC~\cite{hu2022semantic} & ResNet-18+ResNet-50 & 71.2 & 68.3 & 74.0 & 61.9 & 83.0 & 38.4 \\
        \midrule
        \textbf{SMTC} & ViT-S/16 & 64.0 & 61.8 & 66.3 & 45.7 & 75.5 & 38.1 \\
        \textbf{SMTC$^\dagger$} & ViT-S/16 & 67.6 & 65.2 & 69.9 & 53.2 & 79.6 & \textbf{39.2} \\
        \textbf{SMTC} & ResNet-50 & \textbf{73.0} & \textbf{69.4} & \textbf{76.6} & \textbf{62.5} & \textbf{84.1} & 38.8 \\
    \bottomrule
    \end{tabular}
    \caption{Quantitative results on label propagation tasks: semi-supervised video object segmentation on DAVIS-2017, pose tracking on JHMDB, human part tracking on VIP. We present the backbone for comprehensive comparison, and report comparing results from~\cite{xu2021rethinking,hu2022semantic}. And the method denoted with $^\dagger$ denotes operating label propagation with the downsample ratio of 8.}
    \label{propagation}
\end{table*}

\noindent\textbf{Single Object Discovery.}
To measure our model's ability to decompose different objects, we first present the quantitative results on single object discovery without post processing (e.g., CRF, spectral clustering) in Table~\ref{discover}. Many existing works resort to optical flow~\cite{yang2019unsupervised,lamdouar2021segmenting,yang2021self,xie2022segmenting} or synthetic data~\cite{xie2022segmenting} as weak supervision to learn temporal dynamics and generate semantic-agnostic segmentation masks for moving objects. DINO baseline~\cite{caron2021emerging} uses the self-attention map from the last layer as an object segmentation mask. While our method only relies on RGB frames and explicitly discriminates object semantics in a fully self-supervised manner. From the comparison, we observe that our SMTC with only semantic decomposition largely exceeds RGB-only methods~\cite{caron2021emerging,liu2021emergence} and comparable to optical flow-based methods, and our formulation with instance identification further improves the performance. This phenomenon indicates that the learned semantic decomposition with temporal correspondence cues have the potential to substitute pre-computed optical flow to guide single object segmentation.

\begin{figure}
    \centering
    \includegraphics[width=\linewidth]{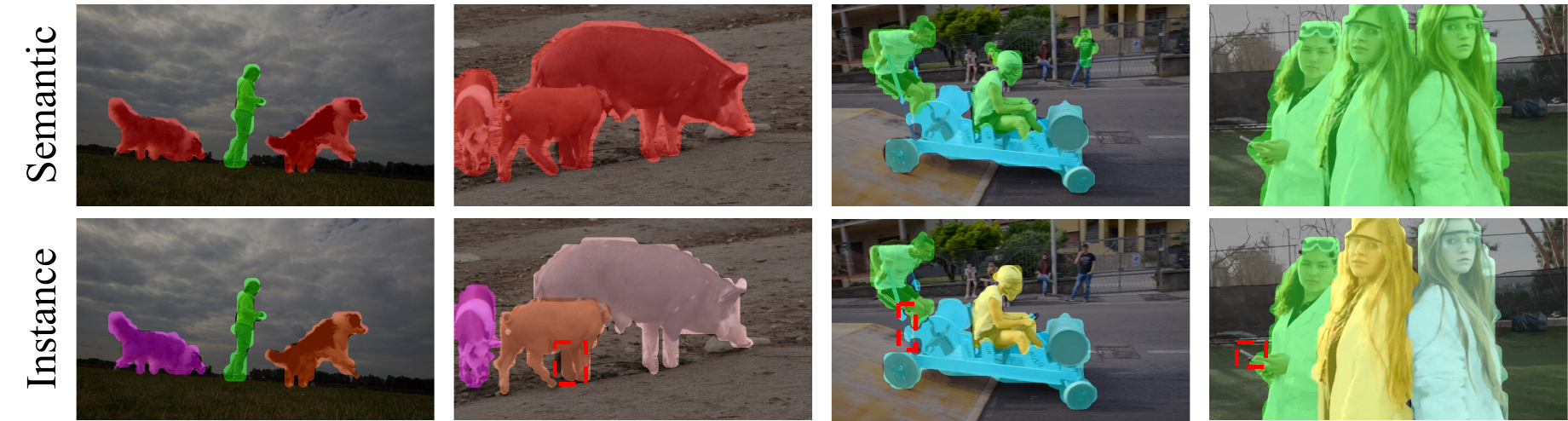}
    \caption{Visualization of semantic and instance segmentation map. The \textcolor{red}{red} boxes outline the ambiguous areas.}
    \label{fig:multiple}
\end{figure}

\noindent\textbf{Multiple Object Discovery.}
Most of the existing video segmentation methods with fully unsupervised training cannot explicitly distinguish multiple object instances. OCLR~\cite{xie2022segmenting} adopts layered flow representations to identify multiple objects on re-annotated DAVIS-2017-motion dataset, but it cannot distinguish instances with common motion. On the contrary, our method jointly utilizes semantics and temporal correspondence to separate arbitrary object instances on DAVIS-2017-Unsupervised benchmark. For fair comparison, we re-run DINOSAUR~\cite{seitzer2022bridging} on DAVIS-2017 with $5$ slots as an unsupervised baseline. As presented in Table~\ref{multiple}, our method significantly outperforms DINOSAUR. This is because our method explicitly decomposes different semantics, exploits temporal correspondence among multiple frames to distinguish instances and encourages instance-level temporal consistency. While DINOSAUR is only supervised with DINO feature reconstruction without considering temporal relationships and semantic discrimination. Besides, our method is comparable with supervised baseline RVOS~\cite{ventura2019rvos}, and the gap to supervised state-of-the-art~\cite{lin2021video} is majorly due to human annotated masks as supervision leading to much more precise segmentation boundaries. This conjecture can be demonstrated by Fig.~\ref{fig:multiple}, which also reveals that our model is able to learn discriminative object-centric representations to separate different instances.

\noindent\textbf{Label Propagation.}
Finally, we compare the performance on label propagation tasks in Table~\ref{propagation} to validate the temporal consistency of our learned features. Note that here the performance of DINO on DAVIS-2017 is higher than that reported in the original paper~\cite{caron2021emerging}. To illustrate it, we take ViT-S/16 backbone as an example. Since some images in DAVIS-2017 have dimensions that are not multiples of 16, thus there is information loss in the patch embedding process. To this end, in our evaluation, we apply a post processing to guarantee that the dimensions of all images are resized to multiples of 16. Besides, for comparison with CNN based method with downsample ratio of 8, we conduct an additional experiments by interpolating the features from ViT-S/16 to the same size as the CNNs, and perform label propagation on the interpolated feature maps. The results are denoted with $^\dagger$ in Table~\ref{propagation}.
Though not specifically designed for these tasks, our model achieves state-of-the-art results on three benchmarks. The most related work to our SMTC is SFC~\cite{hu2022semantic}, which also utilizes both high-level semantic and low-level correspondence. The difference is that SFC only employs late fusion to incorporate these two cues in inference, while our model jointly exploits semantics and temporal correspondence in training to identify object instances, perform instance-level alignment and refine temporally consistent object-centric representations.

\subsection{Ablation Study}
We report IoU on DAVIS-2016, and $\mathcal{J}\&\mathcal{F}$ on DAVIS-2017-Unsupervised for single and multiple object discovery. Refer to Supplementary Material for more ablations.
 
\begin{table}[]
    \centering
    \begin{tabular}{ccccccc}
    \toprule
        Model & $N$ & Semantic & Masked & IoU & $\mathcal{J}\&\mathcal{F}$ \\
        \midrule
        Ours-A & 16 & \checkmark & \checkmark & 71.8 & 40.5 \\
         \midrule
        Ours-B & 16 & \checkmark & \xmark & 66.4 & 24.5 \\
        Ours-C & 16 & \xmark & \checkmark & 52.1 & 30.3 \\
         \midrule
        Ours-D & 1 & - & \checkmark & 47.4 & 22.1 \\
        Ours-E & 8 & \checkmark & \checkmark & 68.4 & 37.9 \\
        Ours-F & 32 & \checkmark & \checkmark & 71.9 & 40.7 \\
         \bottomrule
    \end{tabular}
    \caption{Ablation studies on the number of learnable Gaussian distributions, and the formulation of semantic-aware masked slot attention. `Semantic' (`Instance') denotes the first (second) attention stage for semantic decomposition (instance identification).}
    \label{tab:ablation}
\end{table}

\noindent\textbf{Slot Attention Formulation.}
We compare the different formulations of semantic-aware masked slot attention in Table~\ref{tab:ablation}. With only the first slot attention stage, i.e., the `Semantic' part, the model cannot distinguish object instances of the same semantics, thus the performance on multiple object discovery drops significantly as indicated by Ours-B. While with only the second stage, i.e., the `Instance' part, the performance on both single and multiple object segmentation drops as indicated by Ours-C. This reveals that the pre-computed semantic masks play an important role in alleviating interference in instance slot update and filtering valid sample for instance-level alignment.

\noindent\textbf{Number of Learnable Gaussian Distributions.} Comparing Ours-A, Ours-E and Ours-F in Table~\ref{tab:ablation}, we observe that the performance improves when $N$ increases since larger $N$ guides our model to distill more fine-grained semantics. And we present an extreme setting with $N=1$ in Ours-D, where our formulation degenerates into the original slot attention with random sampling from one Gaussian distribution~\cite{locatello2020object}. Note that although Ours-C also has no access to semantic masks, it still maintains the potential to discriminate different semantics due to multiple learnable means. Its superiority to Ours-D demonstrates the effectiveness of semantic discrimination in object discovery.

\begin{table}[]
    \centering
    \begin{tabular}{ccccc}
    \toprule
        Feature & Semantic & Instance & IoU & $\mathcal{J}\&\mathcal{F}$ \\
        \midrule
        Fused & \checkmark & \checkmark & 71.8 & 40.5 \\
        \midrule
        \multirow{3}{*}{\shortstack{RGB-\\only}} & \checkmark & \checkmark & 67.7 & 20.1 \\
         & \checkmark & \xmark & 67.5 & 18.8 \\
         & \xmark & \checkmark & 41.5 & 15.3 \\
        \midrule
        \multirow{3}{*}{\shortstack{Correlation-\\only}} & \checkmark & \checkmark & 38.8 & 14.2 \\
         & \checkmark & \xmark & 21.1 & 10.7 \\
         & \xmark & \checkmark & 59.4 & 27.5 \\
        \bottomrule
    \end{tabular}
    \caption{Ablation studies on the feature usage. We compare different slot attention formulations on different feature inputs.}
    \label{tab:feature}
\end{table}

\noindent\textbf{Feature Usage.} We explore using different feature input to our semantic-aware masked slot attention in Table~\ref{tab:feature}. We observe that with RGB-only input, the `Semantic' part guarantees satisfactory performance on single object discovery, but performs poor on multiple object benchmarks even with the `Instance' part. This coincides with our intuition that it is difficult to distinguish object instances with solely semantic cues, instead we need to look into more frames and resort to temporal correspondence. While for the correlation-only input, the `Instance' part coarsely segments objects and the `Semantic' part exerts negative impact. This is because the correspondence map contains little semantic information, the semantic decomposition results in severe ambiguity, thus impairing the performance.

\begin{table}
    \centering
    \begin{tabular}{ccccc}
    \hline
        $\mathcal{L}_{sem}$ & $\mathcal{L}_{reg}$ & $\mathcal{L}_{obj}$ & IoU & $\mathcal{J}\&\mathcal{F}$  \\
        \hline
        \checkmark & \xmark & \xmark & 61.3 & 21.6 \\
        \checkmark & \checkmark & \xmark & 66.4 & 24.5 \\
        \xmark & \xmark & \checkmark & 50.6 & 30.4 \\
        \checkmark & \checkmark & \checkmark & 71.7 & 40.5 \\
        \hline
    \end{tabular}
    \caption{Ablation studies on the learning objectives. We report the results on DAVIS-2016 and DAVIS-2017-Unsuperivsed.}
    \label{loss}    
\end{table}

\noindent\textbf{Learning Objectives.} We also present a comprehensive ablation on the learning objectives in Table~\ref{loss}, where we report single object discovery on DAVIS-2016 and multiple object discovery results on DAVIS-2017-Unsupervised. From the comparison, we observe that $\mathcal{L}_{sem}$ is fundamental to decompose object semantics, which demonstrates the necessity of semantic decomposition in the first stage. $\mathcal{L}_{reg}$ also leads to improvement by encouraging the learnable queries to cover more diverse semantics. As for $\mathcal{L}_{obj}$, this objective introduces instance-level alignment and significantly enhances multiple object discovery performance on DAVIS-2017-Unsupervised.

\begin{table}
    \centering
    \begin{tabular}{cccc}
    \hline
        T & FPS & IoU & $\mathcal{J}\&\mathcal{F}$  \\
        \hline
        2 & 4 & 65.4 & 36.9 \\
        4 & 2 & 66.3 & 37.7 \\
        4 & 4 & 71.8 & 40.5 \\
        8 & 4 & 71.9 & 40.8 \\
        \hline
    \end{tabular}
    \caption{Ablation studies on the hyper-parameters of frame sampling. We compare different number of frames per clip and different FPS.}
    \label{sample}    
\end{table}

\noindent\textbf{Frame Sampling.} Another important thing in training is the frame sampling procedure. The number of frames per clip and sampling frame rate determines the temporal reception field of our model. We compare different sampling hyper-parameters in Table~\ref{sample}. The results indicate that it is necessary to use large FPS to provide rich temporal dynamics. And by setting comparatively large FPS, our model can have access to more temporal information without introducing more frames, reaching a trade-off between performance and efficiency.



\begin{figure}
    \centering
    \includegraphics[width=\linewidth]{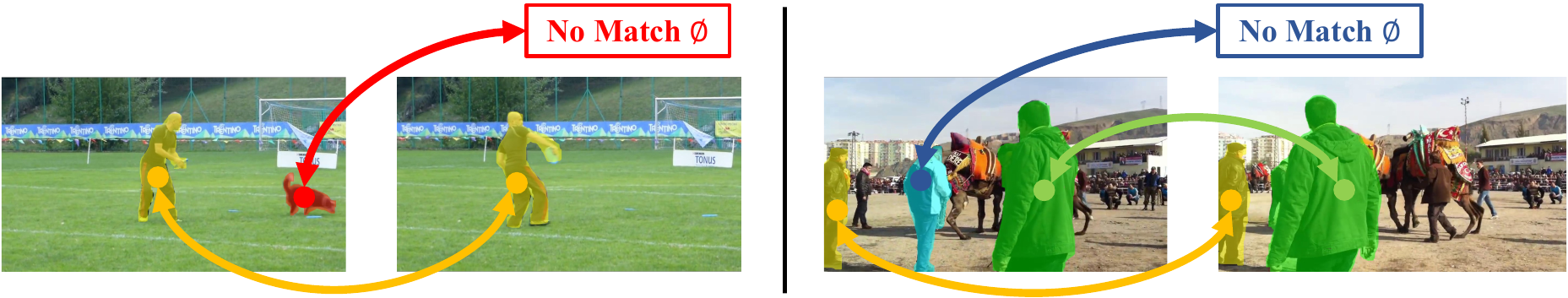}
    \caption{Visualization of instance alignment. The arrows point out the matched instances across time.}
    \label{fig:align}
\end{figure}

\subsection{Further Discussion}
\noindent\textbf{Visualization of Object Discovery.}
We visualize the our semantic decomposition map and instance identification map in Fig.~\ref{fig:multiple}, where the same color denotes the segmentation mask of the same semantics. The visualization reveals that our method is able to distinguish multiple object instances with semantic structure. For example, different people belong to the same semantic center, and quadrupeds such as dogs and pigs belong to another semantics. And our model also separates different instances of the same semantics, such as multiple persons and different pigs, with minor ambiguity on small object part, e.g., pig legs, mobile phones.
Besides, we also visualize our instance-level alignment during training in Fig.~\ref{fig:align}. Our bipartite matching well aligns the corresponding instances in different frames, and the valid sample constraint of Eq.~\ref{criteria} effectively handles object occlusion in videos. For example,  when encountering an occluded person instance, our method filters out the redundant slots not attending to objects, only correlating valid slots for robust instance-level alignment.

\noindent\textbf{Limitation.}
Due to the lack of the human annotated segmentation masks, our model faces challenges in generating precise boundaries for each instance, particularly for small objects. One potential solution is to incorporate multi-scale feature pyramid to further improve dense perception. We leave it in the future work. Despite this limitation, our work effectively demonstrates the benefits of leveraging both semantics and temporal correspondence to discover object instances with semantic structure and distill discriminative and temporally consistent object-centric representations.

\section{Conclusion}
In this work, we propose a novel self-supervised framework jointly exploiting high-level semantics and low-level temporal correspondence to enhance object-centric perception. Specifically, we represent semantic and temporal correspondence cues using the RGB feature map and dense feature correlation, respectively. These cues are fused and fed into semantic-aware masked slot attention which comprises a set of learnable Gaussian distributions. This design allows us to leverage the mean vectors as potential semantic centers for semantic decomposition, and use the perturbations introduced by the standard deviation vectors around the semantic centers to make use of the temporal correspondence cues for instance identification. To distill discriminative and temporally consistent object-centric representations, we devise semantic- and instance-level alignment that is robust to object occlusion as self-supervision. We demonstrate the effectiveness of our model for object-centric analysis through the state-of-the-art performance on label propagation tasks, as well as the promising results on unsupervised object discovery in both single and multiple object scenarios.

\section*{Acknowledgement}
We thank Tianfan Xue for constructive feedback.
This project is funded in part by Shanghai AI Laboratory (P23KS00020, 2022ZD0160201), CUHK Interdisciplinary AI Research Institute, 
and the Centre for Perceptual and Interactive Intelligence (CPII) Ltd under the Innovation and Technology Commission (ITC)'s InnoHK.

{\small
\bibliographystyle{ieee_fullname}
\bibliography{egbib}

\begin{thebibliography}{10}\itemsep=-1pt

\bibitem{alwassel2020self}
Humam Alwassel, Dhruv Mahajan, Bruno Korbar, Lorenzo Torresani, Bernard Ghanem,
  and Du Tran.
\newblock Self-supervised learning by cross-modal audio-video clustering.
\newblock {\em Advances in Neural Information Processing Systems},
  33:9758--9770, 2020.

\bibitem{asano2019self}
Yuki~Markus Asano, Christian Rupprecht, and Andrea Vedaldi.
\newblock Self-labelling via simultaneous clustering and representation
  learning.
\newblock {\em arXiv preprint arXiv:1911.05371}, 2019.

\bibitem{bachman2019learning}
Philip Bachman, R~Devon Hjelm, and William Buchwalter.
\newblock Learning representations by maximizing mutual information across
  views.
\newblock {\em Advances in neural information processing systems}, 32, 2019.

\bibitem{baek2020psynet}
Kyungjune Baek, Minhyun Lee, and Hyunjung Shim.
\newblock Psynet: Self-supervised approach to object localization using point
  symmetric transformation.
\newblock In {\em Proceedings of the AAAI Conference on Artificial
  Intelligence}, volume~34, pages 10451--10459, 2020.

\bibitem{besbinar2021self}
Beril Besbinar and Pascal Frossard.
\newblock Self-supervision by prediction for object discovery in videos.
\newblock In {\em 2021 IEEE International Conference on Image Processing
  (ICIP)}, pages 1509--1513. IEEE, 2021.

\bibitem{bian2022learning}
Zhangxing Bian, Allan Jabri, Alexei~A Efros, and Andrew Owens.
\newblock Learning pixel trajectories with multiscale contrastive random walks.
\newblock In {\em Proceedings of the IEEE/CVF Conference on Computer Vision and
  Pattern Recognition}, pages 6508--6519, 2022.

\bibitem{burgess2019monet}
Christopher~P Burgess, Loic Matthey, Nicholas Watters, Rishabh Kabra, Irina
  Higgins, Matt Botvinick, and Alexander Lerchner.
\newblock Monet: Unsupervised scene decomposition and representation.
\newblock {\em arXiv preprint arXiv:1901.11390}, 2019.

\bibitem{caelles20192019}
Sergi Caelles, Jordi Pont-Tuset, Federico Perazzi, Alberto Montes,
  Kevis-Kokitsi Maninis, and Luc Van~Gool.
\newblock The 2019 davis challenge on vos: Unsupervised multi-object
  segmentation.
\newblock {\em arXiv preprint arXiv:1905.00737}, 2019.

\bibitem{carion2020end}
Nicolas Carion, Francisco Massa, Gabriel Synnaeve, Nicolas Usunier, Alexander
  Kirillov, and Sergey Zagoruyko.
\newblock End-to-end object detection with transformers.
\newblock In {\em Computer Vision--ECCV 2020: 16th European Conference,
  Glasgow, UK, August 23--28, 2020, Proceedings, Part I 16}, pages 213--229.
  Springer, 2020.

\bibitem{caron2018deep}
Mathilde Caron, Piotr Bojanowski, Armand Joulin, and Matthijs Douze.
\newblock Deep clustering for unsupervised learning of visual features.
\newblock In {\em Proceedings of the European conference on computer vision
  (ECCV)}, pages 132--149, 2018.

\bibitem{caron2020unsupervised}
Mathilde Caron, Ishan Misra, Julien Mairal, Priya Goyal, Piotr Bojanowski, and
  Armand Joulin.
\newblock Unsupervised learning of visual features by contrasting cluster
  assignments.
\newblock {\em Advances in Neural Information Processing Systems},
  33:9912--9924, 2020.

\bibitem{caron2021emerging}
Mathilde Caron, Hugo Touvron, Ishan Misra, Herv{\'e} J{\'e}gou, Julien Mairal,
  Piotr Bojanowski, and Armand Joulin.
\newblock Emerging properties in self-supervised vision transformers.
\newblock In {\em Proceedings of the IEEE/CVF International Conference on
  Computer Vision}, pages 9650--9660, 2021.

\bibitem{carreira2017quo}
Joao Carreira and Andrew Zisserman.
\newblock Quo vadis, action recognition? a new model and the kinetics dataset.
\newblock In {\em proceedings of the IEEE Conference on Computer Vision and
  Pattern Recognition}, pages 6299--6308, 2017.

\bibitem{chen2020simple}
Ting Chen, Simon Kornblith, Mohammad Norouzi, and Geoffrey Hinton.
\newblock A simple framework for contrastive learning of visual
  representations.
\newblock In {\em International conference on machine learning}, pages
  1597--1607. PMLR, 2020.

\bibitem{cheng2021per}
Bowen Cheng, Alex Schwing, and Alexander Kirillov.
\newblock Per-pixel classification is not all you need for semantic
  segmentation.
\newblock {\em Advances in Neural Information Processing Systems},
  34:17864--17875, 2021.

\bibitem{choudhury2022guess}
Subhabrata Choudhury, Laurynas Karazija, Iro Laina, Andrea Vedaldi, and
  Christian Rupprecht.
\newblock Guess what moves: unsupervised video and image segmentation by
  anticipating motion.
\newblock {\em arXiv preprint arXiv:2205.07844}, 2022.

\bibitem{crawford2020exploiting}
Eric Crawford and Joelle Pineau.
\newblock Exploiting spatial invariance for scalable unsupervised object
  tracking.
\newblock In {\em Proceedings of the AAAI Conference on Artificial
  Intelligence}, volume~34, pages 3684--3692, 2020.

\bibitem{cuturi2013sinkhorn}
Marco Cuturi.
\newblock Sinkhorn distances: Lightspeed computation of optimal transport.
\newblock {\em Advances in neural information processing systems}, 26, 2013.

\bibitem{ding2022motion}
Shuangrui Ding, Maomao Li, Tianyu Yang, Rui Qian, Haohang Xu, Qingyi Chen, Jue
  Wang, and Hongkai Xiong.
\newblock Motion-aware contrastive video representation learning via
  foreground-background merging.
\newblock In {\em Proceedings of the IEEE/CVF Conference on Computer Vision and
  Pattern Recognition}, pages 9716--9726, 2022.

\bibitem{ding2022dual}
Shuangrui Ding, Rui Qian, and Hongkai Xiong.
\newblock Dual contrastive learning for spatio-temporal representation.
\newblock In {\em Proceedings of the 30th ACM International Conference on
  Multimedia}, pages 5649--5658, 2022.

\bibitem{ding2022mod}
Shuangrui Ding, Weidi Xie, Yabo Chen, Rui Qian, Xiaopeng Zhang, Hongkai Xiong,
  and Qi Tian.
\newblock Motion-inductive self-supervised object discovery in videos.
\newblock {\em arXiv preprint arXiv:2210.00221}, 2022.

\bibitem{ding2023prune}
Shuangrui Ding, Peisen Zhao, Xiaopeng Zhang, Rui Qian, Hongkai Xiong, and Qi
  Tian.
\newblock Prune spatio-temporal tokens by semantic-aware temporal accumulation.
\newblock {\em arXiv preprint arXiv:2308.04549}, 2023.

\bibitem{dosovitskiy2020image}
Alexey Dosovitskiy, Lucas Beyer, Alexander Kolesnikov, Dirk Weissenborn,
  Xiaohua Zhai, Thomas Unterthiner, Mostafa Dehghani, Matthias Minderer, Georg
  Heigold, Sylvain Gelly, et~al.
\newblock An image is worth 16x16 words: Transformers for image recognition at
  scale.
\newblock {\em arXiv preprint arXiv:2010.11929}, 2020.

\bibitem{dosovitskiy2015flownet}
Alexey Dosovitskiy, Philipp Fischer, Eddy Ilg, Philip Hausser, Caner Hazirbas,
  Vladimir Golkov, Patrick Van Der~Smagt, Daniel Cremers, and Thomas Brox.
\newblock Flownet: Learning optical flow with convolutional networks.
\newblock In {\em Proceedings of the IEEE international conference on computer
  vision}, pages 2758--2766, 2015.

\bibitem{elsayed2022savi++}
Gamaleldin~F Elsayed, Aravindh Mahendran, Sjoerd van Steenkiste, Klaus Greff,
  Michael~C Mozer, and Thomas Kipf.
\newblock Savi++: Towards end-to-end object-centric learning from real-world
  videos.
\newblock {\em arXiv preprint arXiv:2206.07764}, 2022.

\bibitem{emami2021efficient}
Patrick Emami, Pan He, Sanjay Ranka, and Anand Rangarajan.
\newblock Efficient iterative amortized inference for learning symmetric and
  disentangled multi-object representations.
\newblock In {\em International Conference on Machine Learning}, pages
  2970--2981. PMLR, 2021.

\bibitem{engelcke2019genesis}
Martin Engelcke, Adam~R Kosiorek, Oiwi~Parker Jones, and Ingmar Posner.
\newblock Genesis: Generative scene inference and sampling with object-centric
  latent representations.
\newblock {\em arXiv preprint arXiv:1907.13052}, 2019.

\bibitem{engelcke2021genesis}
Martin Engelcke, Oiwi Parker~Jones, and Ingmar Posner.
\newblock Genesis-v2: Inferring unordered object representations without
  iterative refinement.
\newblock {\em Advances in Neural Information Processing Systems},
  34:8085--8094, 2021.

\bibitem{feichtenhofer2021large}
Christoph Feichtenhofer, Haoqi Fan, Bo Xiong, Ross Girshick, and Kaiming He.
\newblock A large-scale study on unsupervised spatiotemporal representation
  learning.
\newblock In {\em Proceedings of the IEEE/CVF Conference on Computer Vision and
  Pattern Recognition}, pages 3299--3309, 2021.

\bibitem{gidaris2018unsupervised}
Spyros Gidaris, Praveer Singh, and Nikos Komodakis.
\newblock Unsupervised representation learning by predicting image rotations.
\newblock {\em arXiv preprint arXiv:1803.07728}, 2018.

\bibitem{girshick2015fast}
Ross Girshick.
\newblock Fast r-cnn.
\newblock In {\em Proceedings of the IEEE international conference on computer
  vision}, pages 1440--1448, 2015.

\bibitem{greff2019multi}
Klaus Greff, Rapha{\"e}l~Lopez Kaufman, Rishabh Kabra, Nick Watters,
  Christopher Burgess, Daniel Zoran, Loic Matthey, Matthew Botvinick, and
  Alexander Lerchner.
\newblock Multi-object representation learning with iterative variational
  inference.
\newblock In {\em International Conference on Machine Learning}, pages
  2424--2433. PMLR, 2019.

\bibitem{grill2020bootstrap}
Jean-Bastien Grill, Florian Strub, Florent Altch{\'e}, Corentin Tallec, Pierre
  Richemond, Elena Buchatskaya, Carl Doersch, Bernardo Avila~Pires, Zhaohan
  Guo, Mohammad Gheshlaghi~Azar, et~al.
\newblock Bootstrap your own latent-a new approach to self-supervised learning.
\newblock {\em Advances in neural information processing systems},
  33:21271--21284, 2020.

\bibitem{han2017scnet}
Kai Han, Rafael~S Rezende, Bumsub Ham, Kwan-Yee~K Wong, Minsu Cho, Cordelia
  Schmid, and Jean Ponce.
\newblock Scnet: Learning semantic correspondence.
\newblock In {\em Proceedings of the IEEE international conference on computer
  vision}, pages 1831--1840, 2017.

\bibitem{han2019video}
Tengda Han, Weidi Xie, and Andrew Zisserman.
\newblock Video representation learning by dense predictive coding.
\newblock In {\em Proceedings of the IEEE/CVF International Conference on
  Computer Vision Workshops}, pages 0--0, 2019.

\bibitem{han2020self}
Tengda Han, Weidi Xie, and Andrew Zisserman.
\newblock Self-supervised co-training for video representation learning.
\newblock {\em Advances in Neural Information Processing Systems},
  33:5679--5690, 2020.

\bibitem{he2022masked}
Kaiming He, Xinlei Chen, Saining Xie, Yanghao Li, Piotr Doll{\'a}r, and Ross
  Girshick.
\newblock Masked autoencoders are scalable vision learners.
\newblock In {\em Proceedings of the IEEE/CVF Conference on Computer Vision and
  Pattern Recognition}, pages 16000--16009, 2022.

\bibitem{he2020momentum}
Kaiming He, Haoqi Fan, Yuxin Wu, Saining Xie, and Ross Girshick.
\newblock Momentum contrast for unsupervised visual representation learning.
\newblock In {\em Proceedings of the IEEE/CVF conference on computer vision and
  pattern recognition}, pages 9729--9738, 2020.

\bibitem{he2017mask}
Kaiming He, Georgia Gkioxari, Piotr Doll{\'a}r, and Ross Girshick.
\newblock Mask r-cnn.
\newblock In {\em Proceedings of the IEEE international conference on computer
  vision}, pages 2961--2969, 2017.

\bibitem{he2016deep}
Kaiming He, Xiangyu Zhang, Shaoqing Ren, and Jian Sun.
\newblock Deep residual learning for image recognition.
\newblock In {\em Proceedings of the IEEE conference on computer vision and
  pattern recognition}, pages 770--778, 2016.

\bibitem{held2016learning}
David Held, Sebastian Thrun, and Silvio Savarese.
\newblock Learning to track at 100 fps with deep regression networks.
\newblock In {\em European conference on computer vision}, pages 749--765.
  Springer, 2016.

\bibitem{henaff2020data}
Olivier Henaff.
\newblock Data-efficient image recognition with contrastive predictive coding.
\newblock In {\em International conference on machine learning}, pages
  4182--4192. PMLR, 2020.

\bibitem{hu2022semantic}
Yingdong Hu, Renhao Wang, Kaifeng Zhang, and Yang Gao.
\newblock Semantic-aware fine-grained correspondence.
\newblock {\em arXiv preprint arXiv:2207.10456}, 2022.

\bibitem{jabri2020space}
Allan Jabri, Andrew Owens, and Alexei Efros.
\newblock Space-time correspondence as a contrastive random walk.
\newblock {\em Advances in neural information processing systems},
  33:19545--19560, 2020.

\bibitem{jhuang2013towards}
Hueihan Jhuang, Juergen Gall, Silvia Zuffi, Cordelia Schmid, and Michael~J
  Black.
\newblock Towards understanding action recognition.
\newblock In {\em Proceedings of the IEEE international conference on computer
  vision}, pages 3192--3199, 2013.

\bibitem{jia2022unsupervised}
Baoxiong Jia, Yu Liu, and Siyuan Huang.
\newblock Unsupervised object-centric learning with bi-level optimized query
  slot attention.
\newblock {\em arXiv preprint arXiv:2210.08990}, 2022.

\bibitem{jing2018self}
Longlong Jing, Xiaodong Yang, Jingen Liu, and Yingli Tian.
\newblock Self-supervised spatiotemporal feature learning via video rotation
  prediction.
\newblock {\em arXiv preprint arXiv:1811.11387}, 2018.

\bibitem{kabra2021simone}
Rishabh Kabra, Daniel Zoran, Goker Erdogan, Loic Matthey, Antonia Creswell,
  Matt Botvinick, Alexander Lerchner, and Chris Burgess.
\newblock Simone: View-invariant, temporally-abstracted object representations
  via unsupervised video decomposition.
\newblock {\em Advances in Neural Information Processing Systems},
  34:20146--20159, 2021.

\bibitem{kim2018learning}
Dahun Kim, Donghyeon Cho, Donggeun Yoo, and In~So Kweon.
\newblock Learning image representations by completing damaged jigsaw puzzles.
\newblock In {\em 2018 IEEE Winter Conference on Applications of Computer
  Vision (WACV)}, pages 793--802. IEEE, 2018.

\bibitem{kipf2021conditional}
Thomas Kipf, Gamaleldin~F Elsayed, Aravindh Mahendran, Austin Stone, Sara
  Sabour, Georg Heigold, Rico Jonschkowski, Alexey Dosovitskiy, and Klaus
  Greff.
\newblock Conditional object-centric learning from video.
\newblock {\em arXiv preprint arXiv:2111.12594}, 2021.

\bibitem{lai2020mast}
Zihang Lai, Erika Lu, and Weidi Xie.
\newblock Mast: A memory-augmented self-supervised tracker.
\newblock In {\em Proceedings of the IEEE/CVF Conference on Computer Vision and
  Pattern Recognition}, pages 6479--6488, 2020.

\bibitem{lai2019self}
Zihang Lai and Weidi Xie.
\newblock Self-supervised learning for video correspondence flow.
\newblock {\em arXiv preprint arXiv:1905.00875}, 2019.

\bibitem{lamdouar2021segmenting}
Hala Lamdouar, Weidi Xie, and Andrew Zisserman.
\newblock Segmenting invisible moving objects.
\newblock 2021.

\bibitem{li2013video}
Fuxin Li, Taeyoung Kim, Ahmad Humayun, David Tsai, and James~M Rehg.
\newblock Video segmentation by tracking many figure-ground segments.
\newblock In {\em Proceedings of the IEEE international conference on computer
  vision}, pages 2192--2199, 2013.

\bibitem{li2019joint}
Xueting Li, Sifei Liu, Shalini De~Mello, Xiaolong Wang, Jan Kautz, and
  Ming-Hsuan Yang.
\newblock Joint-task self-supervised learning for temporal correspondence.
\newblock {\em Advances in Neural Information Processing Systems}, 32, 2019.

\bibitem{lin2021video}
Huaijia Lin, Ruizheng Wu, Shu Liu, Jiangbo Lu, and Jiaya Jia.
\newblock Video instance segmentation with a propose-reduce paradigm.
\newblock In {\em Proceedings of the IEEE/CVF International Conference on
  Computer Vision}, pages 1739--1748, 2021.

\bibitem{liu2021emergence}
Runtao Liu, Zhirong Wu, Stella Yu, and Stephen Lin.
\newblock The emergence of objectness: Learning zero-shot segmentation from
  videos.
\newblock {\em Advances in Neural Information Processing Systems},
  34:13137--13152, 2021.

\bibitem{liu2020self}
Songtao Liu, Zeming Li, and Jian Sun.
\newblock Self-emd: Self-supervised object detection without imagenet.
\newblock {\em arXiv preprint arXiv:2011.13677}, 2020.

\bibitem{locatello2020object}
Francesco Locatello, Dirk Weissenborn, Thomas Unterthiner, Aravindh Mahendran,
  Georg Heigold, Jakob Uszkoreit, Alexey Dosovitskiy, and Thomas Kipf.
\newblock Object-centric learning with slot attention.
\newblock {\em Advances in Neural Information Processing Systems},
  33:11525--11538, 2020.

\bibitem{long2015fully}
Jonathan Long, Evan Shelhamer, and Trevor Darrell.
\newblock Fully convolutional networks for semantic segmentation.
\newblock In {\em Proceedings of the IEEE conference on computer vision and
  pattern recognition}, pages 3431--3440, 2015.

\bibitem{loshchilov2018decoupled}
Ilya Loshchilov and Frank Hutter.
\newblock Decoupled weight decay regularization.
\newblock In {\em International Conference on Learning Representations}, 2018.

\bibitem{misra2016shuffle}
Ishan Misra, C~Lawrence Zitnick, and Martial Hebert.
\newblock Shuffle and learn: unsupervised learning using temporal order
  verification.
\newblock In {\em European conference on computer vision}, pages 527--544.
  Springer, 2016.

\bibitem{ng2018actionflownet}
Joe Yue-Hei Ng, Jonghyun Choi, Jan Neumann, and Larry~S Davis.
\newblock Actionflownet: Learning motion representation for action recognition.
\newblock In {\em 2018 IEEE Winter Conference on Applications of Computer
  Vision (WACV)}, pages 1616--1624. IEEE, 2018.

\bibitem{ochs2013segmentation}
Peter Ochs, Jitendra Malik, and Thomas Brox.
\newblock Segmentation of moving objects by long term video analysis.
\newblock {\em IEEE transactions on pattern analysis and machine intelligence},
  36(6):1187--1200, 2013.

\bibitem{oord2018representation}
Aaron van~den Oord, Yazhe Li, and Oriol Vinyals.
\newblock Representation learning with contrastive predictive coding.
\newblock {\em arXiv preprint arXiv:1807.03748}, 2018.

\bibitem{perazzi2016benchmark}
Federico Perazzi, Jordi Pont-Tuset, Brian McWilliams, Luc Van~Gool, Markus
  Gross, and Alexander Sorkine-Hornung.
\newblock A benchmark dataset and evaluation methodology for video object
  segmentation.
\newblock In {\em Proceedings of the IEEE conference on computer vision and
  pattern recognition}, pages 724--732, 2016.

\bibitem{pont20172017}
Jordi Pont-Tuset, Federico Perazzi, Sergi Caelles, Pablo Arbel{\'a}ez, Alex
  Sorkine-Hornung, and Luc Van~Gool.
\newblock The 2017 davis challenge on video object segmentation.
\newblock {\em arXiv preprint arXiv:1704.00675}, 2017.

\bibitem{qian2022static}
Rui Qian, Shuangrui Ding, Xian Liu, and Dahua Lin.
\newblock Static and dynamic concepts for self-supervised video representation
  learning.
\newblock In {\em European Conference on Computer Vision}, pages 145--164.
  Springer, 2022.

\bibitem{qian2021enhancing}
Rui Qian, Yuxi Li, Huabin Liu, John See, Shuangrui Ding, Xian Liu, Dian Li, and
  Weiyao Lin.
\newblock Enhancing self-supervised video representation learning via
  multi-level feature optimization.
\newblock In {\em Proceedings of the IEEE/CVF international conference on
  computer vision}, pages 7990--8001, 2021.

\bibitem{qian2021spatiotemporal}
Rui Qian, Tianjian Meng, Boqing Gong, Ming-Hsuan Yang, Huisheng Wang, Serge
  Belongie, and Yin Cui.
\newblock Spatiotemporal contrastive video representation learning.
\newblock In {\em Proceedings of the IEEE/CVF Conference on Computer Vision and
  Pattern Recognition}, pages 6964--6974, 2021.

\bibitem{ren2015faster}
Shaoqing Ren, Kaiming He, Ross Girshick, and Jian Sun.
\newblock Faster r-cnn: Towards real-time object detection with region proposal
  networks.
\newblock {\em Advances in neural information processing systems}, 28, 2015.

\bibitem{seitzer2022bridging}
Maximilian Seitzer, Max Horn, Andrii Zadaianchuk, Dominik Zietlow, Tianjun
  Xiao, Carl-Johann Simon-Gabriel, Tong He, Zheng Zhang, Bernhard
  Sch{\"o}lkopf, Thomas Brox, et~al.
\newblock Bridging the gap to real-world object-centric learning.
\newblock {\em arXiv preprint arXiv:2209.14860}, 2022.

\bibitem{sun2018pwc}
Deqing Sun, Xiaodong Yang, Ming-Yu Liu, and Jan Kautz.
\newblock Pwc-net: Cnns for optical flow using pyramid, warping, and cost
  volume.
\newblock In {\em Proceedings of the IEEE conference on computer vision and
  pattern recognition}, pages 8934--8943, 2018.

\bibitem{teed2020raft}
Zachary Teed and Jia Deng.
\newblock Raft: Recurrent all-pairs field transforms for optical flow.
\newblock In {\em Computer Vision--ECCV 2020: 16th European Conference,
  Glasgow, UK, August 23--28, 2020, Proceedings, Part II 16}, pages 402--419.
  Springer, 2020.

\bibitem{valmadre2017end}
Jack Valmadre, Luca Bertinetto, Joao Henriques, Andrea Vedaldi, and Philip~HS
  Torr.
\newblock End-to-end representation learning for correlation filter based
  tracking.
\newblock In {\em Proceedings of the IEEE conference on computer vision and
  pattern recognition}, pages 2805--2813, 2017.

\bibitem{ventura2019rvos}
Carles Ventura, Miriam Bellver, Andreu Girbau, Amaia Salvador, Ferran Marques,
  and Xavier Giro-i Nieto.
\newblock Rvos: End-to-end recurrent network for video object segmentation.
\newblock In {\em Proceedings of the IEEE/CVF conference on computer vision and
  pattern recognition}, pages 5277--5286, 2019.

\bibitem{voigtlaender2017online}
Paul Voigtlaender and Bastian Leibe.
\newblock Online adaptation of convolutional neural networks for video object
  segmentation.
\newblock {\em arXiv preprint arXiv:1706.09364}, 2017.

\bibitem{vondrick2018tracking}
Carl Vondrick, Abhinav Shrivastava, Alireza Fathi, Sergio Guadarrama, and Kevin
  Murphy.
\newblock Tracking emerges by colorizing videos.
\newblock In {\em Proceedings of the European conference on computer vision
  (ECCV)}, pages 391--408, 2018.

\bibitem{wang2013learning}
Naiyan Wang and Dit-Yan Yeung.
\newblock Learning a deep compact image representation for visual tracking.
\newblock {\em Advances in neural information processing systems}, 26, 2013.

\bibitem{wang2017transitive}
Xiaolong Wang, Kaiming He, and Abhinav Gupta.
\newblock Transitive invariance for self-supervised visual representation
  learning.
\newblock In {\em Proceedings of the IEEE international conference on computer
  vision}, pages 1329--1338, 2017.

\bibitem{wang2019learning}
Xiaolong Wang, Allan Jabri, and Alexei~A Efros.
\newblock Learning correspondence from the cycle-consistency of time.
\newblock In {\em Proceedings of the IEEE/CVF Conference on Computer Vision and
  Pattern Recognition}, pages 2566--2576, 2019.

\bibitem{wang2021dense}
Xinlong Wang, Rufeng Zhang, Chunhua Shen, Tao Kong, and Lei Li.
\newblock Dense contrastive learning for self-supervised visual pre-training.
\newblock In {\em Proceedings of the IEEE/CVF Conference on Computer Vision and
  Pattern Recognition}, pages 3024--3033, 2021.

\bibitem{wu2018unsupervised}
Zhirong Wu, Yuanjun Xiong, Stella~X Yu, and Dahua Lin.
\newblock Unsupervised feature learning via non-parametric instance
  discrimination.
\newblock In {\em Proceedings of the IEEE conference on computer vision and
  pattern recognition}, pages 3733--3742, 2018.

\bibitem{xie2022segmenting}
Junyu Xie, Weidi Xie, and Andrew Zisserman.
\newblock Segmenting moving objects via an object-centric layered
  representation.
\newblock {\em arXiv preprint arXiv:2207.02206}, 2022.

\bibitem{xie2021propagate}
Zhenda Xie, Yutong Lin, Zheng Zhang, Yue Cao, Stephen Lin, and Han Hu.
\newblock Propagate yourself: Exploring pixel-level consistency for
  unsupervised visual representation learning.
\newblock In {\em Proceedings of the IEEE/CVF Conference on Computer Vision and
  Pattern Recognition}, pages 16684--16693, 2021.

\bibitem{xu2021rethinking}
Jiarui Xu and Xiaolong Wang.
\newblock Rethinking self-supervised correspondence learning: A video
  frame-level similarity perspective.
\newblock In {\em Proceedings of the IEEE/CVF International Conference on
  Computer Vision}, pages 10075--10085, 2021.

\bibitem{xu2018youtube}
Ning Xu, Linjie Yang, Yuchen Fan, Jianchao Yang, Dingcheng Yue, Yuchen Liang,
  Brian Price, Scott Cohen, and Thomas Huang.
\newblock Youtube-vos: Sequence-to-sequence video object segmentation.
\newblock In {\em Proceedings of the European conference on computer vision
  (ECCV)}, pages 585--601, 2018.

\bibitem{yang2021self}
Charig Yang, Hala Lamdouar, Erika Lu, Andrew Zisserman, and Weidi Xie.
\newblock Self-supervised video object segmentation by motion grouping.
\newblock In {\em Proceedings of the IEEE/CVF International Conference on
  Computer Vision}, pages 7177--7188, 2021.

\bibitem{yang2019unsupervised}
Yanchao Yang, Antonio Loquercio, Davide Scaramuzza, and Stefano Soatto.
\newblock Unsupervised moving object detection via contextual information
  separation.
\newblock In {\em Proceedings of the IEEE/CVF Conference on Computer Vision and
  Pattern Recognition}, pages 879--888, 2019.

\bibitem{zhou2018adaptive}
Qixian Zhou, Xiaodan Liang, Ke Gong, and Liang Lin.
\newblock Adaptive temporal encoding network for video instance-level human
  parsing.
\newblock In {\em Proceedings of the 26th ACM international conference on
  Multimedia}, pages 1527--1535, 2018.

\bibitem{zhu2017flow}
Xizhou Zhu, Yujie Wang, Jifeng Dai, Lu Yuan, and Yichen Wei.
\newblock Flow-guided feature aggregation for video object detection.
\newblock In {\em Proceedings of the IEEE international conference on computer
  vision}, pages 408--417, 2017.

\end{thebibliography}
}

\end{document}